\documentclass[letterpaper, 10 pt, conference]{ieeeconf}  

\IEEEoverridecommandlockouts                              

\usepackage{booktabs}       
\usepackage{amsfonts}       
\usepackage{microtype}      
\usepackage{epsfig}
\usepackage{graphicx}
\usepackage{subfigure} 
\usepackage{floatrow}
\title{\LARGE \bf
Go Wider: An Efficient Neural Network for Point Cloud Analysis via Group Convolutions
}

\author{Can Chen$^{1}$,  Luca Zanotti Fragonara$^{2}$ and Antonios Tsourdos$^{3}$}%

\begin{document}

\maketitle
\thispagestyle{empty}
\pagestyle{empty}

\begin{abstract}

In order to achieve better performance for point cloud analysis, many researchers apply deeper neural networks using stacked Multi-Layer-Perceptron (MLP) convolutions over irregular point cloud. However, applying dense MLP convolutions over large amount of points (e.g. autonomous driving application) leads to inefficiency in memory and computation. To achieve high performance but less complexity, we propose a deep-wide neural network, called ShufflePointNet, to exploit fine-grained local features and reduce redundancy in parallel using group convolution and channel shuffle operation. Unlike conventional operation that directly applies MLPs on high-dimensional features of point cloud, our model goes wider by splitting features into groups in advance, and each group with certain smaller depth is only responsible for respective MLP operation, which can reduce complexity and allows to encode more useful information. Meanwhile, we connect communication between groups by shuffling groups in feature channel to capture fine-grained features. We claim that, multi-branch method for wider neural networks is also beneficial to feature extraction for point cloud.  We present extensive experiments for shape classification task on ModelNet40 dataset and semantic segmentation task on large scale datasets  ShapeNet part, S3DIS and KITTI. We further perform ablation study and compare our model to other state-of-the-art algorithms in terms of complexity and accuracy.

\end{abstract}

\section{INTRODUCTION}

Processing point cloud is increasingly becoming an essential task for a wide range of applications, such as environment perception \cite{zhou2018voxelnet, qi2018frustum, ku2018joint, liu2018real} for autonomous driving, virtual reconstruction \cite{bruder2014poster}, augmented reality (AR). However, it is still challenging to analyze underlying shape representation efficiently due to the disadvantages of large amount of points and unstructured distribution, although point cloud is capable to provide sufficient and accurate geometric information.

Considering remarkable success and advantages of CNNs, \cite{maturana2015voxnet, wang2015voting} apply CNNs over standard grid structure voxelized from unordered point cloud, as CNNs only work on regular grid data. However, this intuitive way leads to high memory and computation due to its naturally sparse and irregular structure. \textit{PointNet} \cite{qi2017pointnet} treats point cloud as a set of unordered points directly, and leverages Multi-Layer-Perceptron (MLP) network and a symmetric function (e.g. max pooling) to exploit global features and make unordered points invariant to permutations. The drawback is that local information is not involved in the stacked MLP layers. For this problem, \textit{PointNet++} \cite{qi2017pointnet++} builds a hierarchical neural network that joints \textit{PointNet} and a sampling and grouping layer to capture local representation. \textit{DGCNN} \cite{wang2018dynamic} extract local features by introducing an edge convolution operation on points and edges connecting each point and corresponding neighbors. \cite{chen2019gapnet} constructs an attention-aware neural network to learn local features by highlighting different attention coefficient for neighboring points.  \textit{PointCNN} \cite{li2018pointcnn} manage to transform an unordered point cloud to a latent canonical order by learning a \(\chi\)-convolutional operator. \cite{liu2019relation} attempts to learn irregular CNN-like filters to capture local features for point cloud. 

\begin{figure}[t!]
  \centering
   \subfigure{\label{fig:acc_time}\includegraphics[width=0.8\linewidth]{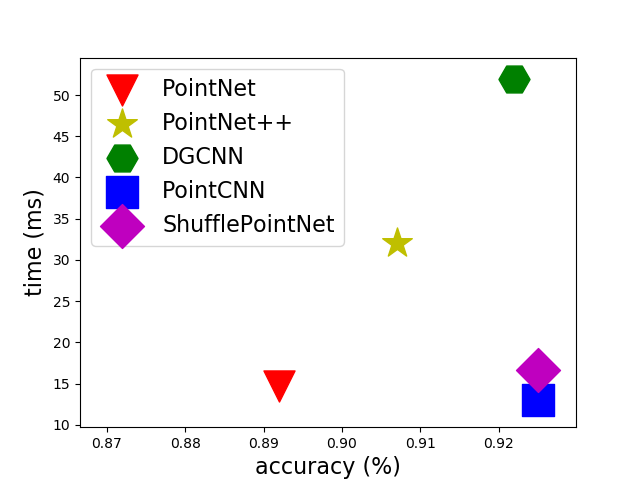}} \vskip -14pt
   \subfigure{\label{fig:acc_flops}\includegraphics[width=0.8\linewidth]{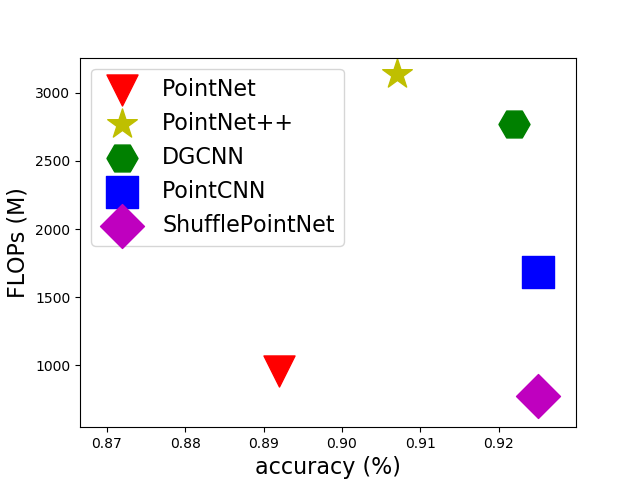}} \vskip -14pt
   \subfigure{\label{fig:acc_params}\includegraphics[width=0.8\linewidth]{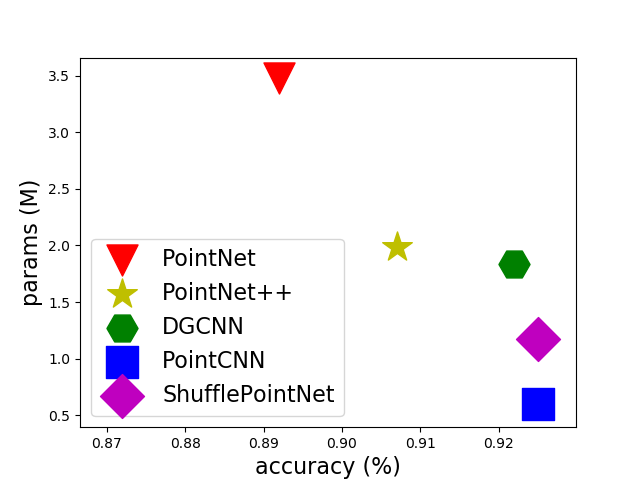}} \vskip -14pt
  \caption{Measurement of accuracy, parameters, FLOPs and forward time. It shows that our proposed ShufflePointNet (magenta diamond marker) always stays on the lower right region, which represents the efficiency of our model in both accuracy and model complexity.}
  \label{fig:meas}
 \end{figure}


We notice that modern neural networks for point cloud \cite{qi2017pointnet, qi2017pointnet++, wang2018dynamic, chen2019gapnet} heavily rely on dense Multi-Layer-Perceptron (MLP) to build repeated block structures with the different number of filters, but eliminating redundancy for dense MLP operation is rarely mentioned for point cloud analysis, although it is common in computer vision domain \cite{xie2017aggregated, howard2017mobilenets, sandler2018mobilenetv2, zhang2018shufflenet, zhang2017interleaved}. In point cloud domain, less redundancy brings more potential for some applications (e.g. autonomous driving) that need to process large scale point cloud. On one hand, the MLP operation is not spatial convolution, which limits the ability of local feature extraction. As a result, applying too many and deeper MLP convolutions easily leads to overfitting. On the other hand, MLP operation is efficient to deal with unstructured data (e.g. point cloud, social networks), as regular CNNs are only available on standard grid structure. Therefore, we draw attention to sparse MLP convolution over irregular point cloud, which can not only leverage advantages of MLP convolutions, but also reduce redundancy by making MLP convolution sparse.

Inspired by group convolution that is introduced by \cite{krizhevsky2012imagenet, zhang2018shufflenet} and can be treated as standard convolution with sparse filters, we primarily focus on building a deep-wide neural network to reduce complexity and also achieve high performance. The main contributions of this paper are summarized as follows:
\begin{itemize}
\item An efficient deep-wide neural network, named \textit{ShufflePointNet}, is introduced, aimed at significantly reducing redundancy in depth and exploiting more feature information in width for point cloud understanding.
\item We combine advantages of both MLP and group convolution to achieve better performance with less complexity for point cloud analysis.
\item We propose to split features into groups, and to use the shuffle operation to exchange information between groups.
\item We evaluate our model on extensive benchmarks, including small point cloud dataset (e.g. ModelNet40, ShapeNet part), large-scale point cloud (e.g. S3DIS) and super large-scale point cloud (KITTI), to represent efficiency over different scales data.
\item To the best of our knowledge, ShufflePointNet is the first deep-wide model employing channel shuffle operation and group convolution to efficiently capture local representations for point cloud. 
\end{itemize}

\begin{figure*}[h]
  \centering
   {\epsfig{file = 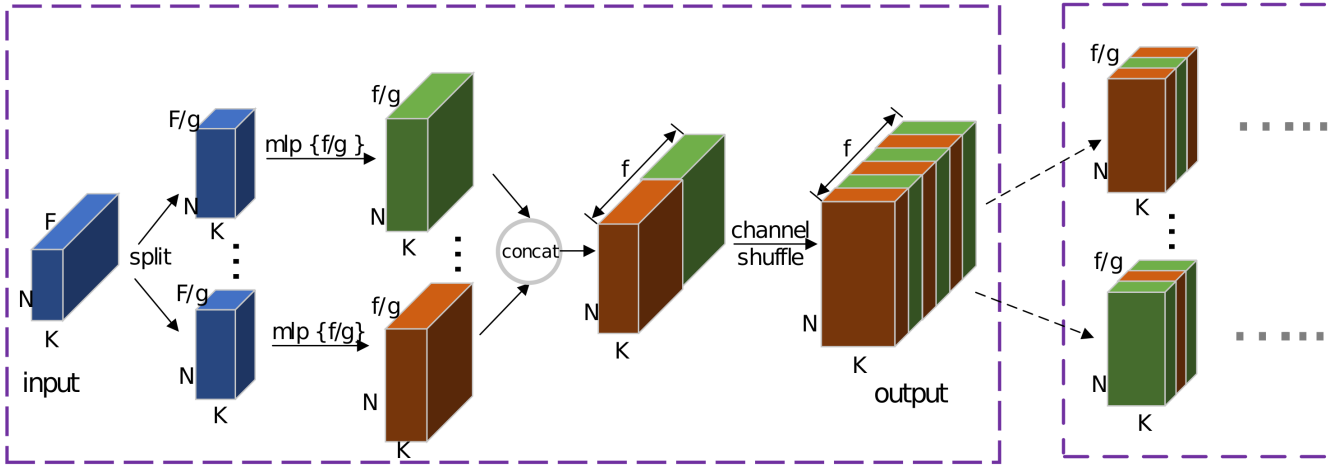, width=1\linewidth}}
  \caption{\textbf{MLP Shuffled Group Convolution unit (SGC unit).} We take the feature with \(N \times K \times F\) as input, where \(N, K, F\) indicate the number of points, neighbors and feature channels respectively. Assuming the dimension of output feature is \(N \times K \times f\). For the standard MLP convolution, we apply \(1 \times 1\) convolution with \(f\) filters to input feature, and the number of parameters is \(F \times f\). For our SGC unit, the number of parameters is reduced to \(\frac{F \times f}{g}\), where \(g\) is the number of groups.}
  \label{fig:sgc}
 \end{figure*}

\section{Related work}

\paragraph{Volumetric grid and multi-view methods.} Volumetric methods \cite{maturana2015voxnet, klokov2017escape, bentley1975multidimensional, riegler2017octnet} convert irregular point cloud to regular dense grid and voxels to allow feature extraction by standard CNNs. However, applying CNNs over dense volumetric grid leads to extremely high cost on computation and memory. \cite{klokov2017escape, riegler2017octnet} improve efficiency in space partition and resolution, but still cause loss of local geometric information due to bounding voxels.

Multi-view methods \cite{qi2016volumetric, wang2017dominant} apply classic 2D CNNs on group of 2D image views that are obtained from 3D objects by different angles. However, 3D geometric shape is unlikely to be captured from these 2D images due to lack of depth information. Besides, part of points information might be lost due to the occlusions on the images. As a result, it is non-trivial to segment every point for classification.

\paragraph{Deep learning methods directly on unstructured point cloud.} \textit{PointNet} \cite{qi2017pointnet} takes the lead in treating point cloud as a set of unordered points and applying stacked Multi-Layer-Perceptron (MLP) to learn individual point feature. The global feature is finally obtained by a symmetric function (e.g. max pooling). However, local region features that are beneficial to better understanding geometric shape are not considered in this approach. In order to improve performance and remedy this drawback, \textit{PointNet++} \cite{qi2017pointnet++} constructs a hierarchical neural network that recursively applies \textit{PointNet} on local features that combine sampled points with corresponding neighboring points. \textit{DGCNN} \cite{wang2018dynamic} applies \textit{PointNet} on edge features that concatenate each point and its edges connecting corresponding point and its neighboring pairs. \textit{PointCNN} \cite{li2018pointcnn} transforms a given irregular point cloud to a latent canonical order by learning a \(\chi\)-convolutional operator, after which classic 2D CNNs are available to use for local feature extraction. \cite{liu2019relation} attempts to learn a customized convolutional weight from geometric local relation for shape-aware representation.

\paragraph{Geometric deep learning methods.} Geometric deep learning \cite{bronstein2017geometric} is defined as a modern term for deep neural network techniques that address non-Euclidean structured data (e.g. point cloud, social networks). Graph CNNs \cite{bruna2013spectral, defferrard2016convolutional, zhang2018graph} have achieved great success in many tasks for graph representation of non-Euclidean data. \textit{Superpoint} \cite{landrieu2018large} organizes point set into geometric elements, after which a graph CNN structure is applied to exploit local features.

\section{Model architecture}
In this section, we explain the architecture of our model resorting to two components: MLP Shuffled Group Convolution unit (SGC unit for short) as shown in Figure~\ref{fig:sgc} and model architecture in Figure~\ref{fig:model}. We define \(X=\left\{ x_i \in \mathbb{R}^F, i=1,2,\ldots,N\right\}\) as a raw point set and input for our model, where \(F\) is the dimension of point representation,  \(N\) is the number of points, and \(x_i\) is 3D geometric position of each point. Other observations,such as color, intensity, surface normal, can also be used to augment each point feature information.

\subsection{Local feature representation} Representing point cloud to graph structure with nodes and edges is an applicable method due to the fact that each node and corresponding edges on a graph can be naturally defined point cloud as each point feature and its neighborhood. As a result, converting point cloud to a graph and applying neural networks on the graph structure is efficient to learn embedding information for neighborhood of each node. 

We then construct a directed graph \(G=(V,E)\) for a point set, where \(V \subseteq\ \mathbb{R}^F\) is node for each point, and \(E\) indicates the corresponding edges to neighborhood. Assuming the input of point set is more or less uniform distribution, we choose \(k\)-nearest neighbor (\(k\)-NN) search to explore neighborhood of each point, as it can guarantee fixed number of neighbors. We define \(N_i\) as a neighborhood set of each point \(x_i\), then feature vector of directed edge is defined as \({e_i}_j=(x_i, x_i-{x_{i}}_j)\), where center point \(x_i \in V\), neighboring point \(x_j \in N_i\), and \({x_i}_j\) is \(x_j\) to \(x_i\).

\subsection{Model complexity metrics} In order to show efficiency of group convolution, we firstly introduce several metrics to measure model complexity, such as memory size, computational cost and forward time. Specifically, memory size is calculated by the total number of parameters for convolution kernels, the number of floatingpoint operations (FLOPs) indicates computational cost of neural networks, forward time represents forward propagation time of neural networks. FLOPs is a widely used but indirect metric, and it can approximate model complexity by counting part of operations in the neural networks, such as the number of multiply-adds operation. The forward time is normally used as direct metric to measure model speed. It is worth mentioning that FLOPs and forward time vary on the different experimental environment. Besides, forward time is a necessary metric to measure model speed, as FLOPs is not sufficient to indicate all the operations in the model and corresponding time consuming.

\subsection{Group convolution} \textit{AlexNet} \cite{krizhevsky2012imagenet} first introduces group convolution for distributing the model over two GPUs. Its effectiveness is also well demonstrated by \cite{xie2017aggregated, howard2017mobilenets, sandler2018mobilenetv2, zhang2018shufflenet, zhang2017interleaved} in image domain. We study that group convolution has several representative advantages. Firstly, it can ease the training of deep neural networks by reducing redundancy; Secondly, it is an efficient method to increase the width of the neural networks that allows more feature channels which is beneficial to encoding more information on each group if we consider constrained complexity budget; Thirdly, it can be also treated as sparse convolution, as convolution kernels on each group are only responsible to part of entire feature channel. Last but not least, it can relieve overfitting to some extent.

\begin{figure*}[t!]
  \centering
   {\epsfig{file = 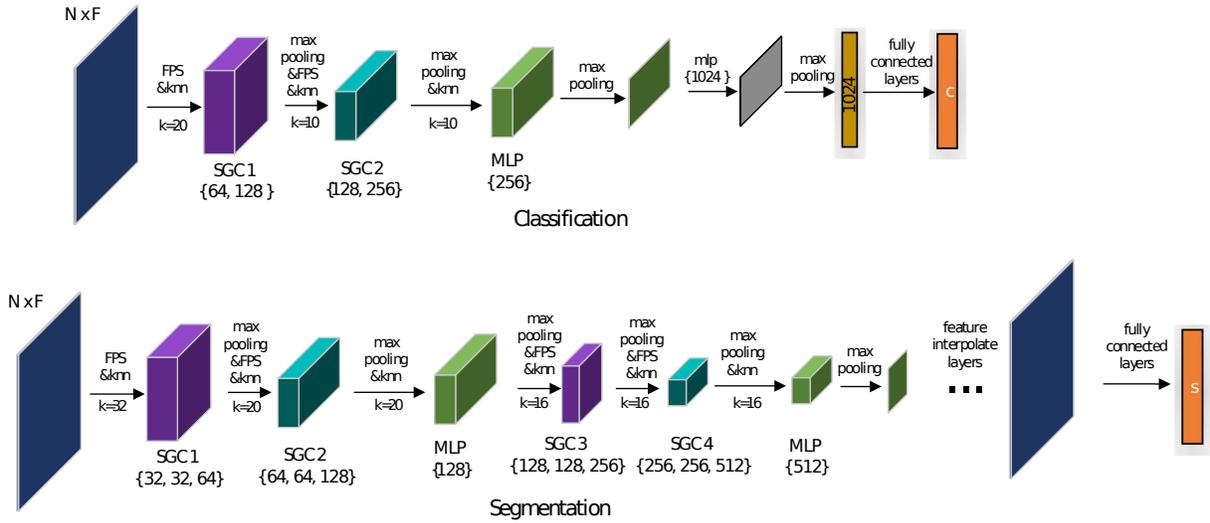, width=0.9\linewidth}}
  \caption{\textbf{ShufflePointNet architecture:} The architecture use \textit{PointNet++} as backbone, and it contains classification structure (top branch) and semantic segmentation structure (bottom branch) for point cloud. In detail, \(N, F\) indicates the number of input points, and corresponding feature channels respectively. \(k\) is the number of the neighbors. Besides, the number in the brace \{\} indicates the MLP filters.}
  \label{fig:model}
 \end{figure*}

The memory size and computational complexity (FLOPs) for a single \(1 \times 1\) group convolution can be calculated by Equations \ref{eq:memory} and \ref{eq:flops} ~\cite{ma2018shufflenet} respectively. It is easy to observe that it becomes to standard MLP operation when we set \(g=1\). Besides, the number of parameters and FLOPs are reduced to \(1/g\) for a single operation when we split \(k\)-nn graph to \(g\) groups.

\begin{equation}\label{eq:memory}
    params=1 \times 1 \times \frac{{C_{in}}}{g} \times \frac{{C_{out}}}{g} \times g = \frac{{C_{in}} \times {C_{out}}}{g}
\end{equation}
\begin{equation}\label{eq:flops}
    FLOPs=\frac{N \times K \times {C_{in}} \times {C_{out}}}{g}
\end{equation}

where \({C_{in}}\) and \({C_{out}}\) indicate input channel and output channel respectively, \(N\) is the number of points, \(K\) is the number of neighboring points, \(g\) is the number of groups.

\subsection{Channel shuffle} Considering the fact that each group only holds incomplete and part representations of graph features, it is unlikely to capture sufficient features if there is no communication among multiple group convolutions. Inspired by \cite{zhang2018shufflenet}, we stack group convolutions together and then shuffle the feature channel to sufficiently fuse features for all groups. We split fused features into \(g\) groups again for the input of next layer as shown in Figure ~\ref{fig:sgc}. As a result, each update group contains information of all other groups.

\subsection{Model architecture}
Our ShufflePointNet architecture for shape classification and segmentation is shown in Figure ~\ref{fig:model}. We use \textit{PointNet++} \cite{qi2017pointnet++} as our backbone. However, there are three main differences between our model and \textit{PointNet++}. Firstly, we use \(1 \times 1\) group convolutions to exploit local fine-grained features in both depth and width. Secondly, instead of radius search for neighboring points in \textit{PointNet++}, we employ \(k\)-NN search to guarantee fixed number of neighbors. Thirdly, instead of neighboring point feature, edge feature connecting each point and corresponding neighbors is used to explicitly represent the relative position of neighboring point to its center point.

\section{Experiments}
In this section, we perform comprehensive experiments to evaluate our ShufflePointNet model in both shape classification and segmentation tasks. To demonstrate effectiveness of our model, we compare accuracy and complexity to recent state-of-the-art methods. We further arrange ablation study to carefully investigate different settings.

\begin{table*}[t!]
  \caption{Semantic part segmentation results on ShapeNet part dataset.}
  \label{tab:seg_result} \centering
 \begin{tabular}{p{2.3cm}|p{0.6cm}|p{0.4cm}p{0.4cm}p{0.4cm}p{0.4cm}p{0.4cm}p{0.4cm}p{0.4cm}p{0.4cm}p{0.4cm}p{0.4cm}p{0.4cm}p{0.4cm}p{0.4cm}p{0.4cm}p{0.4cm}p{0.4cm}}
    \toprule[0.5pt]
    & avg & air. & bag  & cap  & car  & cha. & ear. & gui. & kni. & lam. & lap. & mot. & mug  & pis. & roc. & ska. & tab. \\ \midrule
kd-net \cite{klokov2017escape}                                                 & 82.3 & 82.3 & 74.6 & 74.3 & 70.3 & 88.6  & 73.5          & 90.2   & 87.2  & 81.0 & 94.9   & 57.4  & 86.7 & 78.1   & 51.8   & 69.9        & 80.3  \\
Kc-Net \cite{shen2018mining}                                              & 83.7 & 82.8 & 81.5 & 86.4 & 77.6 & 90.3  & 76.8    & 91.0   & 87.2  & 84.5 & 95.5   & 69.2  & 94.4 & 81.6   & 60.1  &75.2   &81.3\\ 
PointNet \cite{qi2017pointnet}                                               & 83.7 & 83.4 & 78.7 & 82.5 & 74.9 & 89.6  & 73.0       & \textbf{91.5}   & 85.9  & 80.8 & 95.3   & 65.2  & 93.0 & 81.2   & 57.9   & 72.8       & 80.6  \\
3DmFV  \cite{ben20173d}                                         & 84.3 & 82.0 & 84.3 & 86.0 & 76.9  & 89.9       & 73.9   & 90.8  & 85.7 & 82.6   & 95.2  & 66.0 & 94.0   & 82.6   & 51.5       & 73.5 & 81.8  \\
RSNet   \cite{huang2018recurrent}                                             & 84.9  & 82.7 & \textbf{86.4} & 84.1& \textbf{78.2}  & 90.4       & 69.3   & 91.4  & 87.0 & 83.5   & 95.4  & 66.0 & 92.6   & 81.8   & 56.1    & 75.8  & 82.2 \\
PointNet++  \cite{qi2017pointnet++}                                             & 85.1 & 82.4 & 79.0 & \textbf{87.7} & 77.3 & 90.8  & 71.8         & 91.0   & 85.9  & 83.7 & 95.3   & 71.6  & 94.1 & 81.3   & 58.7   & 76.4               & 82.6  \\
DGCNN  \cite{wang2018dynamic}                                                  & 85.1 & \textbf{84.2} & 83.7 & 84.4 & 77.1 & \textbf{90.9}  & 78.5         & \textbf{91.5}   & 87.3  & 82.9 & \textbf{96.0}   & 67.8  & 93.3 & 82.6   & 59.7   & 75.5         & 82.0  \\ 
SGPN                                                    & \textbf{85.8} & 80.4 &78.6 & 78.8 & 71.5 & 88.6  & 78.0        & 90.9   & 83.0  & 78.8 & 95.8   & \textbf{77.8} & 93.8 & \textbf{87.4}   & \textbf{60.1}  & \textbf{92.3}   & {89.4}\\ \midrule
OURS                                                    & 85.1 & 82.7 &83.6 & 86.2 & 77.3 & 90.3  & \textbf{78.7}          & 90.9   & \textbf{87.4}  & \textbf{84.6} & 94.6   & 68.5  & \textbf{94.5} & 82.9   & 51.3   & 73.4       & 82.3  \\ 
   \bottomrule[0.5pt]
  \end{tabular}
\end{table*}

\subsection{Classification}
\label{cls_point}
\paragraph{Dataset.} We evaluate our classification model on the ModelNet40 benchmark \cite{wu20153d}. It contains 9,843 training models and 2,468 testing models that are classified to 40 classes. We firstly uniformly down sample all models to 1,024 points from total 2,048 points, and then normalize them into the unit sphere. Besides, in order to improve robustness, we further augment the training models by rotating, scaling the sampled points in random, and jittering the position of every point using Gaussian noise with zero mean and 0.01 standard deviation.

\paragraph{Training details.} Our model is implemented using TensorFlow v1.6. Adam optimizer \cite{kingma2014adam} with batch size 32 is employed in the training. The learning rate is initially set to 0.001, and then decays with a rate of 0.7 every 20 epochs to 0.00001. The momentum for batch normalization starts from 0.9 and increases gradually with a decay rate of 0.5 every 20 epochs to 0.99.

\paragraph{Results.} Table~\ref{tab:cls} compares our shape classification model to recent state-of-the-art models for both accuracy and model complexity, including  mean per-class accuracy (mA \%), overall accuracy (OA \%), the number of parameters (Million), FLOPs  (Million) and forward time (ms) on the ModelNet40 benchmark \cite{wu20153d}. Figure~\ref{fig:meas} also shows the efficiency of our model by illustrating that it stays at the lower right region all the time compared to other state-of-the-art models.

It is worth to mention that compared to backbone structure \textit{PointNet++} \cite{qi2017pointnet++}, our model outperforms by 1.8\% for accuracy, and reduced the amount of parameters, FLOPs and forward time significantly by 41\%, 75\%, 48\% respectively. Besides, compared to the complexity of single scale \textit{PointNet++} (params: 1.48M, FLOPs: 1684M, time: 20ms), our model still reduced parameters by 21\%, FLOPs by 54\% and forward time by 17\%. The results convincingly verify the effectiveness of our deep-wide model. 

We observe that \textit{PointCNN} \cite{li2018pointcnn} has larger FLOPs but  least forward time, while \textit{SpecGCN} \cite{wang2018local} has much similar FLOPs to \textit{PointNet} but consumes much longer forward time. It proves our hypothesis that FLOPs is just indirect method to estimate model complexity, and it cannot be used alone to show the efficiency of model speed.

\begin{table}[h]
  \caption{Classification results on ModelNet40 dataset.}
  \label{tab:cls} \centering
  \setlength{\tabcolsep}{0.5mm}
  \begin{tabular}{cccccc}
    \toprule[0.96pt]
    methods  &  mA(\%) & OA(\%) & params & FLOPs & time\\
    \midrule
    VoxNet \cite{maturana2015voxnet}              & 83.0                & 85.9            & -                          & -            & - \\
    PointNet  \cite{qi2017pointnet}                      & 86.0               & 89.2            & 3.48M                  & 957M    & 14.7ms \\
    PointNet++  \cite{qi2017pointnet++}           & -                     & 90.7            & 1.99M                  & 3136M  & 32.0ms    \\
    KC-Net  \cite{shen2018mining}                      & -                     & 91.0            & -                   & -                    & -\\
    SpecGCN  \cite{wang2018local}                      & -                     & 91.5            & 2.05M                   & 1112M                    & 11254ms\\
    KD-Net  \cite{klokov2017escape}                   & -                    & 91.8             & -                        & -             & - \\
    DGCNN  \cite{wang2018dynamic}                  & \textbf{90.2}              & 92.2             & 1.84M             & 2768M    & 52.0ms \\
    PCNN  \cite{atzmon2018point}                       & -              & 92.3             & -             & -    & - \\
    \midrule
    \textbf{Ours}                                                               & 90.1              & 92.5              & 1.17M                  & \textbf{770M}  & 16.7ms  \\ 
    \midrule
    PointCNN  \cite{li2018pointcnn}                     & 88.8              & 92.5             & \textbf{0.6M}             & 1682M      & \textbf{13.0ms}  \\
    RS-CNN  \cite{liu2019relation}                       & -                   & \textbf{93.6}             & -                      & -             & - \\
    \bottomrule[0.96pt]
  \end{tabular}
\end{table}

\paragraph{Ablation study.} We also test our classification model with different settings on the ModelNet40 benchmark \cite{wu20153d}.

Table~\ref{tab:ablation_1} represents the performance for different number of groups.  In order to evenly split feature channels, the initial input of SGC unit shares the same edge feature \({e_i}_j=(x_i, x_i-{x_{i}}_j)\) for all groups. It shows that the results for multiple groups (\(g=2,4\)) perform better than no grouping operation (\(g=1\)), as multiple groups help to capture more information for a given complexity budget. However, much larger group number (e.g. \(g=8\) ) degenerates the performance. We discuss the reason that the useful feature information becomes limited for each group when we split feature channels to too many groups, which leads to the fact that the model is unlikely to learn useful information from individual group.

The number of parameters and FLOPs only have slightly drop, rather than half amount that is mentioned in Equation \ref{eq:memory} and \ref{eq:flops}. The reason is that the fully-connected layers take up much large part of of complexity. However, it still shows effectiveness of group convolution, and our model manage to capture more useful information and reduced by 18.6\% FLOPs compared to \(g=1\).

The forward time also increases when the group number becomes larger, as more groups need more time to store corresponding features and parameters to the cache \cite{ma2018shufflenet}.

Table~\ref{tab:ablation_2} shows different settings for the number of points, low-level edge features \({e_i}_j\) and grouping methods. It indicates that there is no improvement when we add neighbor features to represent local features, and edge features connecting center point to corresponding neighborhood is beneficial to better results. Besides, \(k\)-NN searching slightly outperforms radius searching. We discuss that \(k\)-NN searching is more efficient when the layout of dataset can be more or less treated as uniform distribution.

\begin{table}[h]
  \caption{Model complexity and accuracy for different groups}
  \label{tab:ablation_1} \centering
  \begin{tabular}{ccccc}
    \toprule[0.96pt]
       groups & OA (\%) & params & FLOPs & time \\
    \midrule
    \(g=1\)               & 91.8            & 1.20M                   & 946M     & 15.3ms \\
    \(g=2\)               & 92.5            & 1.17M                  & 770M    & 16.7ms \\
    \(g=4\)               & 92.3            & 1.15M                  & 692M  & 19.5ms    \\
   \(g=8\)               & 91.5            & 1.14M                  & 650M  & 22.7ms    \\
    \bottomrule[0.96pt]
  \end{tabular}
\end{table}

\begin{table}[h]\footnotesize
  \caption{Ablation study of ShufflePointNet.}
  \label{tab:ablation_2} \centering
  \begin{tabular}{c|ccc|c}
    \toprule[0.96pt]
      model & points & edge feature \({e_i}_j\)  & grouping & OA (\%)\\
    \midrule
    A     & 1k          & \((x_i, x_i-{x_{i}}_j)\)                      &knn                 & 92.5  \\
    B     & 1k         & \((x_i, {x_{i}}_j)\)                             &knn                 & 92.0  \\
    C     & 1k         & \((x_i, {x_{i}}_j, x_i-{x_{i}}_j)\)     &knn                  &92.5  \\
    D     & 2k         & \((x_i, x_i-{x_{i}}_j)\)                    &knn                   & 92.7  \\
    E     & 1k        & \((x_i, x_i-{x_{i}}_j)\)                    &radius search     &92.1  \\

    \bottomrule[0.96pt]
  \end{tabular}
\end{table}

\begin{figure*}[h]
  \centering
   \subfigure[Visualization]{\label{fig:objects}\includegraphics[width=0.4\linewidth]{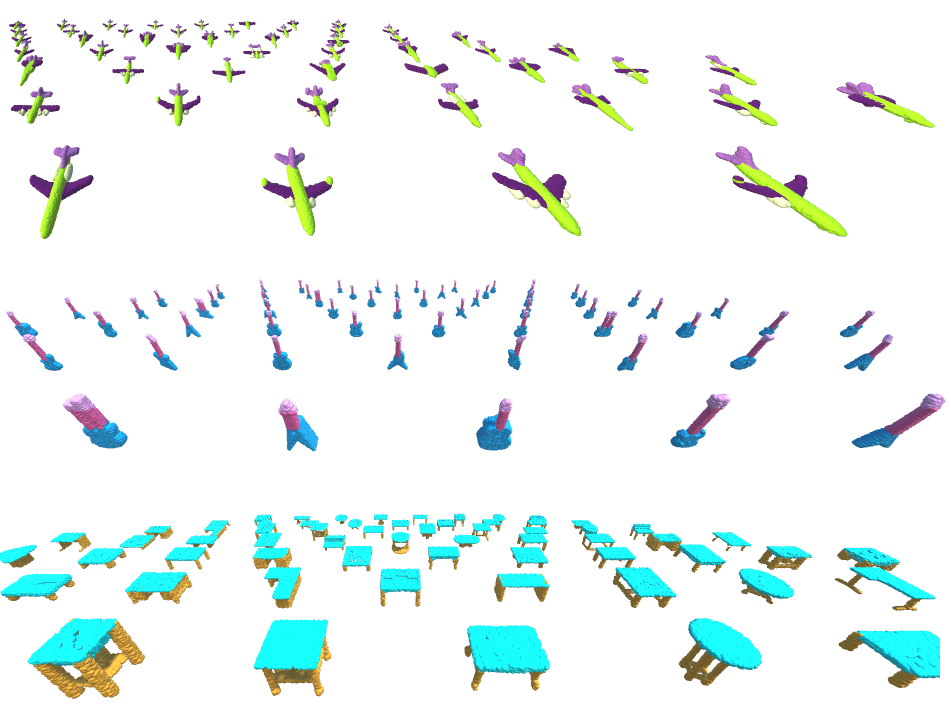}}
   \subfigure[Comparison]{\label{fig:compare}\includegraphics[width=0.3\linewidth]{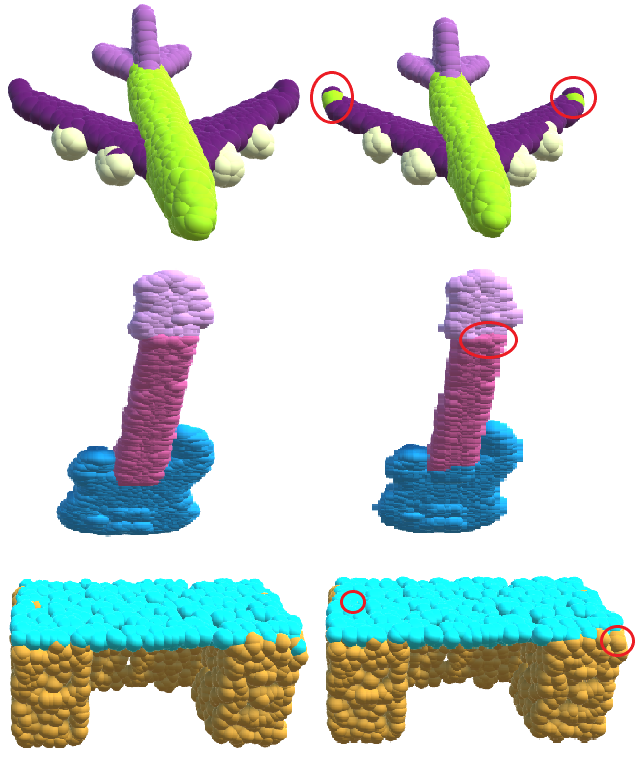}}
  \caption{Visualization of semantic part segmentation results. Figure \subref{fig:objects} visualizes some samples: airplane (top), guitar (middle),  and table (bottom). While Figure \subref{fig:compare} visualizes the predictions and corresponding errors (right) compared to ground truth (left).}
  \label{fig:visual}
 \end{figure*}

\begin{table*}[!]
  \caption{Segmentation results on KITTI.}
  \label{tab:kitti} \centering
  \begin{tabular}{ccccccc}
    \toprule[0.96pt]
     model & mIoU & accuracy & vehicle & cyclist & pedestrian & background\\
    \midrule
    SegCloud \cite{tchapmi2017segcloud}                    & 36.8\%             &-            & 67.5             & 7.3\%             & 53.5\%        & -  \\
    PointNet \cite{qi2017pointnet}                & 38.1\%            & 92\%               & 76.7\%             & 2.9\%               & 6.6\%          & 89.8\% \\
    PCCN \cite{wang2018deep}                    & 58.1\%             & 95.5\%            & \textbf{91.8\%}             & 40.2\%             & 47.7\%        & 89.3\%  \\
    \midrule
    OURS                     & \textbf{72.3\%}   & \textbf{96.8\%}    & 84.9\%   & \textbf{54.0\%}    & \textbf{61.3\%}      & \textbf{95.5\%}    \\

    \bottomrule[0.96pt]
  \end{tabular}
\end{table*}

\subsection{Semantic segmentation}
We evaluate our segmentation model on ShapeNet part dataset \cite{yi2016scalable}, Stanford Large-Scale 3D Indoor Spaces Dataset (S3DIS) \cite{armeni20163d} and KITTI \cite{geiger2013vision}.

\paragraph{ShapeNet part dataset.} The dataset is composed of 16,881 CAD models (14,007 training models and 2,874 testing models ) that are classified to16 categories, and each model is annotated with several parts (less than 6) from 50 part classes. We follow the same sampling strategy as Section~\ref{cls_point} to sample 2,048 points uniformly. The task is to classify each point as part category from models. 

\paragraph{S3DIS dataset.} S3DIS uses Matterport scanners to collect 6-dimension (XYZ, RGB) point clouds, which are then processed to 9D feature (XYZ, RGB, normalized spatial coordinate), from 271 rooms in 6 areas. We follow the settings in \textit{DGCNN} \cite{wang2018dynamic} to slice all the rooms into 1m by 1m blocks, each of which are then sampled to 4,096 points during training process. Meanwhile, we use all points to evaluate our model. We test our model on area 5, and training process is applied on other areas.

\paragraph{KITTI dataset.} We use KITTI Object Detection Benchmark \cite{geiger2013vision} to evaluate our model on real traffic scene. We follow \cite{chen2017multi} to separate the KITTI dataset to 7,481 for training and 191 for testing. However, due to the fact that each frame contains approximate 100,000 points, it is infeasible to apply all points on our model. Therefore, we firstly downsample points from about 100k to 16k (16384 points) by the strategy that we remove points that outside image view, then we randomly select 11,469 points (70\%) in 40 meters and 4915 points for the rest.

\paragraph{Training details.} We follow all the training settings in classification task, except that batch size is set to 24, and we distribute the task to two NVIDIA TESLA V100 GPUs.

\paragraph{Results.} The mean Intersection over Union (mIoU) \cite{qi2017pointnet} is used to evaluate segmentation performance. The IoU is calculated by averaging  IoUs for all parts belonging to the same categories, then the mIoU is the mean IoUs for all shapes from testing dataset.

For the semantic part segmentation task, Table~\ref{tab:seg_result} indicates that our segmentation model achieves competitive results on the ShapeNet part dataset \cite{yi2016scalable}. It wins 4 categories that is the same amount as \textit{DGCNN} and \textit{SGPN}. We also illustrate some shapes from our results in Figure~\ref{fig:objects}, and visualize the errors of our prediction results compared to ground truth as shown in  Figure~\ref{fig:compare} . We represent the ground truth on the left and our predictions and errors on the right.

Table~\ref{tab:s3dis} and ~\ref{tab:kitti} indicate that our segmentation model achieves competitive result on S3DIS and wins the best performance on KITTI dataset. We also observe that both our model and \textit{PointNet++} (reproduced) perform worse for mIoU results, even if the overall accuracy is competitive. We discuss the reason that compared to other models, our model and \textit{PointNet++}  are likely to lose feature information to some extent when apply downsampling and feature interpolation for upsampling layers, although it leads to less model complexity.


\begin{table}[t]
  \caption{Segmentation results on S3DIS Area 5.}
  \label{tab:s3dis} \centering
  \begin{tabular}{ccc}
    \toprule[0.96pt]
     model & mIoU & overall accuracy \\
    \midrule
    SegCloud \cite{tchapmi2017segcloud}                    & 48.9\%             & -    \\
    PointNet \cite{qi2017pointnet}                 & 47.6\%            & 78.5\%       \\
    PointNet++ \cite{qi2017pointnet++} (Repro)                 & 48.3\%            & 84.0\%       \\
    DGCNN \cite{wang2018dynamic}                    & 56.1\%             & 84.1\%    \\
    \textbf{OURS}                     & 50.7\%   & 84.3\%  \\
    PointCNN \cite{li2018pointcnn}                    & 57.3\%             & 85.9\%    \\
    SPGraph \cite{landrieu2018large}                    & \textbf{58.0}\%             & \textbf{86.4\%}    \\
    \bottomrule[0.96pt]
  \end{tabular}
\end{table}

\section{Conclusions}
In this paper, we propose a deep-wide neural network, named ShufflePointNet, to exploit local representations in both depth and width for point cloud. The success of our model verifies that deep-wide neural network can achieve high performance with less model complexity. In the future, we consider to apply our model on the real applications, such as environment perception for our autonomous vehicle that needs to process very large-scale point cloud data.
%


\end{document}